# Accelerating Aquatic Soft Robots with Elastic Instability Effects*

Zechen Xiong[1], Suyu Luohong[2], Jeong Hun Lee[3], Hod Lipson[2]

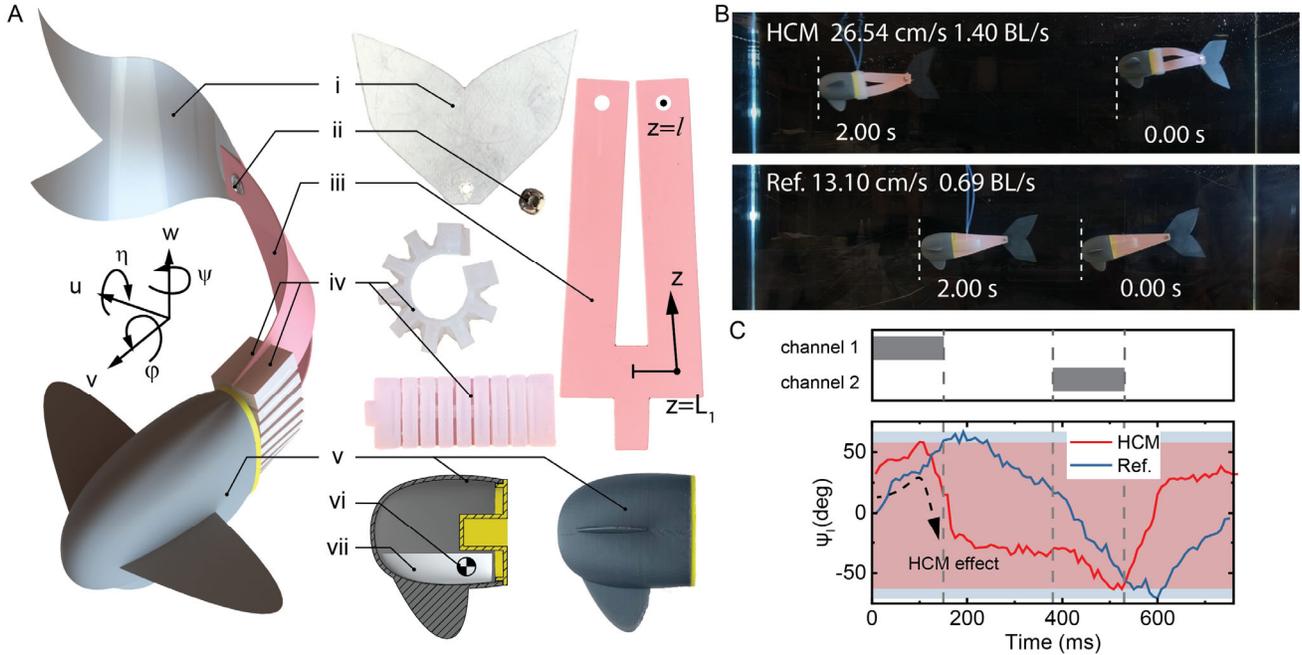

Figure 1 Comparison of the hair-clip-mechanism (HCM) swimming pattern of a pneumatic fish robot and that of a traditional reference. Adapted from [2]. (A) Constitution of the HCM fish robot. i. caudal fin (plastic film with thickness $t$ = 0.191 mm), ii. rivet pin, iii. HCM (plastic film with t = 0.381 mm), iv. antagonistic pneumatic bending units, v. 3D-printed hollow fish head, vi. mass center of the assembly, and vii. cast ballast. (B) Velocity comparison in an aquarium. Scale bar, 150mm. (C) Angular displacement $\psi_l$ w.r.t the forward direction of the two models at the caudal peduncle (rivet pin spot). Both cases use actuation pressure of 150kPa and frequency of 1.3 Hz (period = 760 ms). Grey areas of channel 1 and channel 2 show the duty cycles (150 ms / 380 ms) of actuation in both cases. The red and blue areas show the range of $\psi_l$ in both cases.

*Abstract*—Sinusoidal undulation has long been considered the most successful swimming pattern for fish and bionic aquatic robots [1]. However, a swimming pattern generated by the hair clip mechanism (HCM, part iii, Figure 1A) [2]–[5] may challenge this knowledge. HCM is an in-plane prestressed bi-stable mechanism that stores elastic energy and releases the stored energy quickly via its snap-through buckling. When used for fish robots, the HCM functions as the fish body and creates unique swimming patterns that we term HCM undulation. With the same energy consumption [3], HCM fish outperforms the traditionally designed soft fish with a two-fold increase in cruising speed. We reproduce this phenomenon in a single-link simulation with Aquarium [6]. HCM undulation generates an average propulsion of 16.7 N/m, 2-3 times larger than the reference undulation (6.78 N/m), sine pattern (5.34 N/m/s), and cambering sine pattern (6.36 N/m), and achieves an efficiency close to the sine pattern. These results can aid in developing fish robots and faster swimming patterns.

## I. INTRODUCTION

Fish swimming patterns have long fascinated researchers due to their efficiency, agility, and adaptability in aquatic environments. Related research dates back to the eel locomotion study performed 90 years ago [7]. Modern investigation has shown that 85% of the fish species are Body and/or Caudal Fin (BCF) swimmers [8] that undulate a fraction of their bodies to generate propulsion. Ordered by locomotion speed, BCF swimming is further divided into the diagrams of Ostraciiform, Anguilliform, Subcarangiform, and the fastest Thunniform [9]. These swimming patterns have served as a rich source of inspiration for designing robotic systems that can navigate and interact with water-based environments.

*Research supported by the Dept. of Earth and Environmental Engineering, Columbia University and U.S. National Science Foundation (NSF) AI Institute for Dynamical Systems grant 2112085.

[1]Zechen Xiong is with the Dept. of Earth and Environment Engineering at Columbia University, New York, NY 10027 USA (phone: 9173023864, e-mail: zechen.xiong@columbia.edu).

[2]Suyu Luohong and Hod Lipson is with the Dept. of Mechanical Engineering at Columbia University, New York, NY 10027 USA (e-mail: sl5442@columbia.edu and hod.lipson@columbia.edu).

[3]Jeong Hun Lee is with the Robotics Institute at Carnegie Mellon University, Pittsburgh, PA 15213 USA (e-mail: jeonghunlee@cmu.edu).

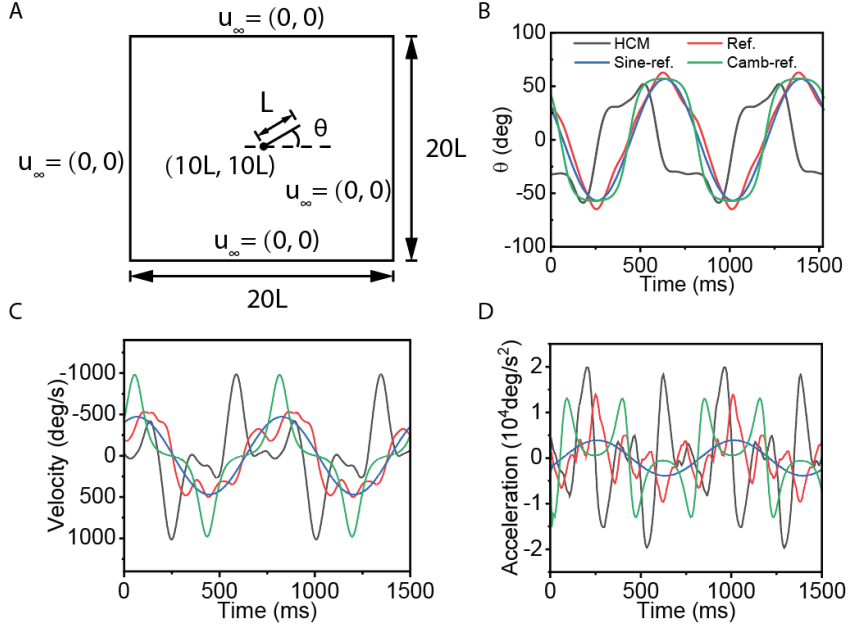

Figure 2 The single-link simulation in Aquarium to reproduce the HCM effect. (A) The setup of the simulation. l = 12 cm, which is the body and caudal fin length of the fish robots. (B) The smoothed HCM undulation, the smoothed reference robot undulation (generated from smoothing plots in Figure 1C), the curve-fit sine wave, and the cambering sine wave as the input signals of the swinging link in the search of the best water-propelling pattern. (C) and (D) The angular velocity and angular acceleration of the four undulation patterns.

On the other hand, BCF swimming is also the major locomotion pattern of soft aquatic robotic systems that use flexible and compliant materials to mimic natural movements and interact with the environment [10]. This new type of fish robot has garnered significant attention due to its potential for safe interaction, economic viability, non-intrusive underwater exploration, and environmental monitoring [11]. Unlike traditional propeller-driven underwater vehicles, soft fish robots usually use fluidic or electroactive polymer actuators to undulate the fish bodies. Marchese et al. [12], Katzschmann et al.[13], Marchese et al. [14], and Katzschmann et al. [11] present the design, fabrication, control, and oceanic testing of a self-contained soft robotic fish capable of rapid and continuum-body motion. The robot swims in three dimensions to monitor aquatic life closely. Li et al. [15] and Li et al. [16] design and fabricate a ray fish-inspired soft swimmer actuated by dielectric elastomer. The swimmer can locomote at high speed and sustain high water pressure in the Mariana Trench. Zhang et al. [17] study global vision-based formation control of soft robotic fish swarm.

While soft robotic fish show great promise, they are plagued by low speeds. The highest speed soft swimmers achieve is 0.5 [11] ~ 0.7 BL/s [15], far from the 2-10 BL/s of organic fish. Some other soft swimmers may have faster speeds but are not adequate for an untethered situation [15], [18], [19]. In our previous studies [3], [4], a type of in-plane prestressed bi-stable hair-clip-like mechanism (HCM, Figure 1A, iii) is proposed to improve the manipulation and locomotion ability of soft robotics. HCMs have elevated stiffness, simple structure, and energy-storing-and-releasing ability. Thus, they can simultaneously function as a structural chassis, motion transmission part, and force amplifier of a robotic system to improve its performance. Our study [1] demonstrates that a pneumatic HCM fish robot can locomote twice as fast as the conventional fish robot (1.40 BL/s versus 0.69 BL/s, Figure 1B) with the same energy input. Observation shows that the HCM body creates a special undulation pattern that we term HCM bi-stable swinging or HCM undulation (Figure 1C).

According to the definition of propulsive efficiency [1]

$$\eta_P = \frac{FU}{P}, \qquad (1)$$

when the swimming speed is doubled, the swimming efficiency can be eight-folded if we assume fluidic resistance is proportional to the velocity squared, i.e.,

$$F = \text{friction} \propto U^2, \qquad (2)$$

where $U$ is the constant forward speed during steady cruising, $P$ is the average input power, and $F$ is the time-averaged force in the forward direction applied on the fish, which is assumed to equal the friction during steady cruising. Studying the unique HCM swimming pattern may provide insights into aquatic soft robotics, efficient underwater vehicles, novel bio-inspired locomotion patterns, etc.

In this work, a comparison of the HCM robot and traditional reference robot (addressed as reference robot, reference, or ref. below) is carried out based on experimental observations; the HCM method, pattern, and analytic solutions are briefly introduced and the aquatic HCM effect is presented in robotics research for the first time, as far as the authors are aware; we verify the HCM effect initially via a simulated single-link undulation and point out the future directions of simulating HCM fish robots.

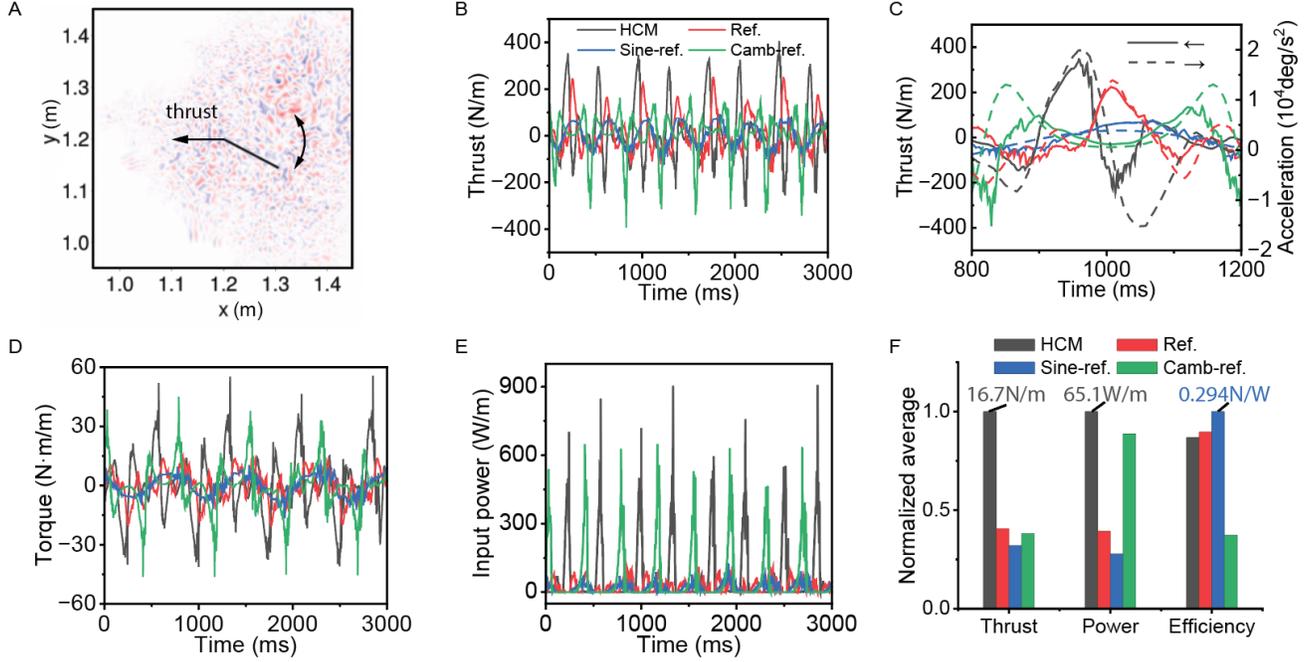

Figure 3 The results of the single-link water-propelling simulation. (A) The link swings to generate a left-ward hydraulic thrust. The color is the vorticity. (B) The thrust from the four difference undulation patterns during a passage of four cycles (3s). (C) The plots of thrust and acceleration indicate a correlation between them two. (D) The torque exerted on the link. (E) The input power of the single-link model, obtained from torque times angular velocity. Negative points are removed to calculate energy efficiencies, since the robots cannot harvest energy. (F)The normalized values of average thrust, average power, and energy efficiencies of the four patterns. HCM undulation gets an average thrust of 16.7 N/m, 2-3 times larger than the reference undulation (6.78 N/m), sine pattern (5.34 N/m), and cambering sine pattern (6.36 N/m), and has a normalized efficiency of 0.87 (0.256 N/W), juxtaposed with efficiencies of 0.90 of the reference wave, 1.00 of the sine wave, and 0.37 of the cambering sine wave.

## II. WORKING PRINCIPLES

### A. HCM swimming pattern

The experiments in [3] are carried out on a pair of pneumatic fish robots with and without HCM (Figure 1). Both use external energy sources, have a pair of antagonistic bending units as the actuation method, and have the same length and self-weight (18.6 cm and 42.5 g). To ensure the same amount of energy consumption, they have the same actuation pressure and duty cycle percentages, as shown in Figure 1C. It is noted that the HCM modulates the near-sinusoidal swinging of the caudal fin into a novel swinging pattern due to its energy-storing and -releasing mechanism. The working principles of the HCM fish robot are described below. Initially, the zero-pressure shape of the HCM is a curved one. When the pneumatic bending unit of the left side (without losing generality) of the fish is active, the HCM starts to build up elastic energy, during which the HCM body and caudal fin bend further to the left. But after a certain pressure level or bending displacement, the HCM releases all its elastic energy accumulated in the previous stage, snapping rapidly to the right and creating an angular speed ~3 times faster (1200 vs 340 °/s) than the reference fish robot. Since the aquatic reaction force is proportional to the velocity squared, HCM undulation ameliorates the swimming efficiency, providing a two-fold speed increase without complicating the robotic design or consuming more energy.

### B. HCM analytic solutions

Our previous modeling, derivation, and verification [3]–[5], [20] have shown that the lateral-torsional buckling accompanies the prestressing process of beams and ribbons. Since the out-of-plane bending moment and rotating torque contribute the most to thin wall beam deformation [20] and the latter is small, we can assume the angled ribbon is straight to simplify the mathematical modeling. With the small deflection assumption, we can depict the angular displacement (Figure 1A) of a cross section on HCM as [3]

$$\varphi(z) = \frac{du}{dz} = \sqrt{l-z} A_1 J_{1/4}\left(\frac{1}{2}\sqrt{\frac{P_{cr}^2}{EI_\eta C}}(l-z)^2\right), \quad (3)$$

in which $z$ is the coordinate along the path of the ribbon (Figure 1A), $u$ or $u(z)$ is the lateral displacement (swaying) of the cross-section, $l$ is the half ribbon length, $A_1$ is a non-zero integration constant that can be determined from energy conservation, $J_{1/4}$ is the Bessel function of the first kind of order ¼, $EI_\eta$ is the out-of-plane bending stiffness of the ribbon, $C$ is the torsional rigidity of the cross-section, $E$ is the Young's modulus, and $P_{cr}$ is the critical load of the lateral-torsional buckling expressed as [3]

$$P_{cr} = \frac{5.5618}{l^2} \cdot \sqrt{EI_\eta C}. \quad (4)$$

From Eq. (1), the lateral displacement of the cross section can be approximated as

$$u(z) = \int_0^z \varphi(s) \, ds \quad (5)$$

The morphing and snapping dynamics of HCM are described in [3].

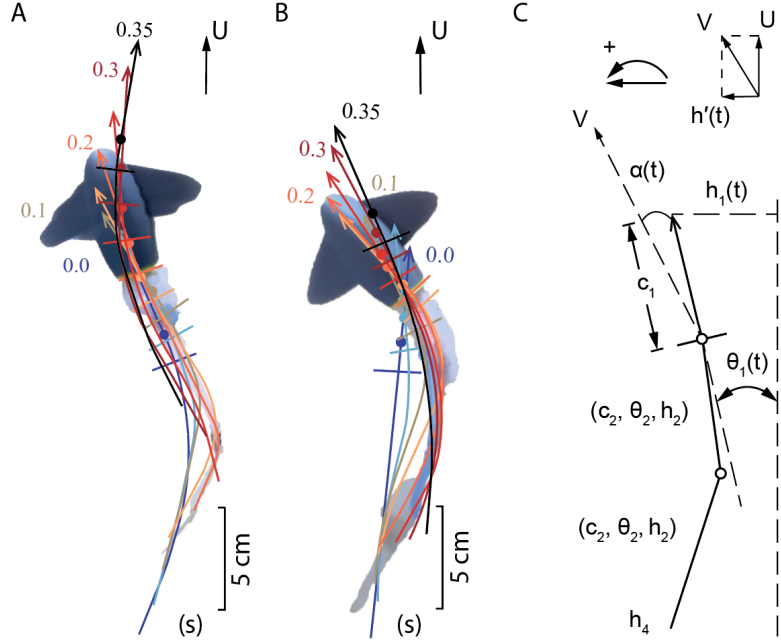

Figure 4 The traced swimming sequences and the sampling methodology for the multi-link simulation. Adapted from [2]. (A) and (B) The robotic configuration evolution of the HCM and reference swimming. (C) The methodology of extracting the three-link model variables from the swimming videos or sequences.

*C. HCM effect simulation*

To study the aquatic HCM effect, we simulate a single-link model of the caudal fin using Aquarium [6]. Aquarium is an open-source, physics-based, fluid-structure interaction solver for robotics that offers stable simulation of the coupled fluid-robot physics in 2D. Specifically, Aquarium simulates the fluid dynamics directly over the normalized Navier-Stokes equations for incompressible, Newtonian fluids (e.g., water). The immersed boundary method [21], [22] is then used to couple the fluid dynamics with a solid body (e.g., BCF), which additionally allows for the calculation of fluid forces acting along the body's surface. The fluid forces acting along the swimming direction are then summed to calculate the net thrust. We refer the reader to the existing literature for more details of Aquarium's implementation [6], [21], [22].

The simulation is set in a 2D box with zero-velocity boundary conditions, as shown in Figure 2A). A swinging rigid link with length $L = 12$ cm is used to replicate the BCF undulation of the fish robots in Figure 1. Figure 2B shows the smoothed peduncle swinging of the HCM fish and the reference from Figure 1C and is used as the simulation input. It is reported in the elongated-body theory [23]–[25] that a traveling wave can describe the cruising of fish

$$h(x,t) = g(x)\sin(kx + \omega t) \quad (6)$$

in which $h(x, t)$ is the lateral displacement from the stretched straight position, $x$ is the coordinate downstream from the nose of the straightened fish, $g(x)$ is the amplitude function, k is the body wave number, and $\omega$ is the body wave frequency. Besides, Xie et al. [25] demonstrate that natural fish use sinusoidal body wave for a higher swimming efficiency, but cambering sinusoidal body wave has higher thrust than sinusoidal one. To broaden the search for the best swimming pattern, we include in the simulation (Figure 2B) a curve-fit sine wave based on the smoothed peduncle undulation of the reference robot and a cambering wave that is expressed as

$$\theta_{\text{cambering}} = \frac{m \tanh(B\theta_{\text{sine}})}{\tanh(B)} \quad (7)$$

where $B > 0$ is the shape control factor, and $\theta_i$ is the undulation angle of the link of the patterns. The cambering wave will have the same amplitude as the sine wave but a higher peak speed, as shown in Figure 2C. Increasing $B$ leads to increasing cambering-ness, and moving $B$ towards zero reduces Eq. (7) to a sine wave. In this work, we take $B = 2$ so that the peak velocity of the cambering swing can match that of the HCM. The HCM undulation has the highest peak acceleration among the four patterns, about 20000 deg/s$^2$ (vs. 14000, 4000, and 13000 deg/s$^2$ for the other three), shown in Figure 2D. A passage of four cycles (~3s, $T = 760$ ms) is simulated.

### III. RESULTS

Figures 3A and 3B show the vorticity and the recorded thrust, respectively. We note that the thrust is positively or negatively correlated with the acceleration in an alternative pattern, as shown by a half period of positive correlation in Figure 3C. This phenomenon can be attributed to the fact that a certain fraction of the fluidic volume at the fish's rear has the same velocity as the swinging fin, which means only the change of velocity, i.e., the acceleration, of the fin pushes against this volume and provides thrust. The alternation is due to the time and geometric symmetry.

Figure 3D is the torque the link experiences during the four types of undulation, and the input power in Figure 3E is obtained by multiplying the torque by the angular velocities in Figure 2C, with negative values removed since the robotic system doesn't collect energy. Integrating the instant thrust and power provides the average thrust, the average power, and the energy efficiencies shown as absolute and normalized values in Figure 3F, with normalization factors being their respective maxima.

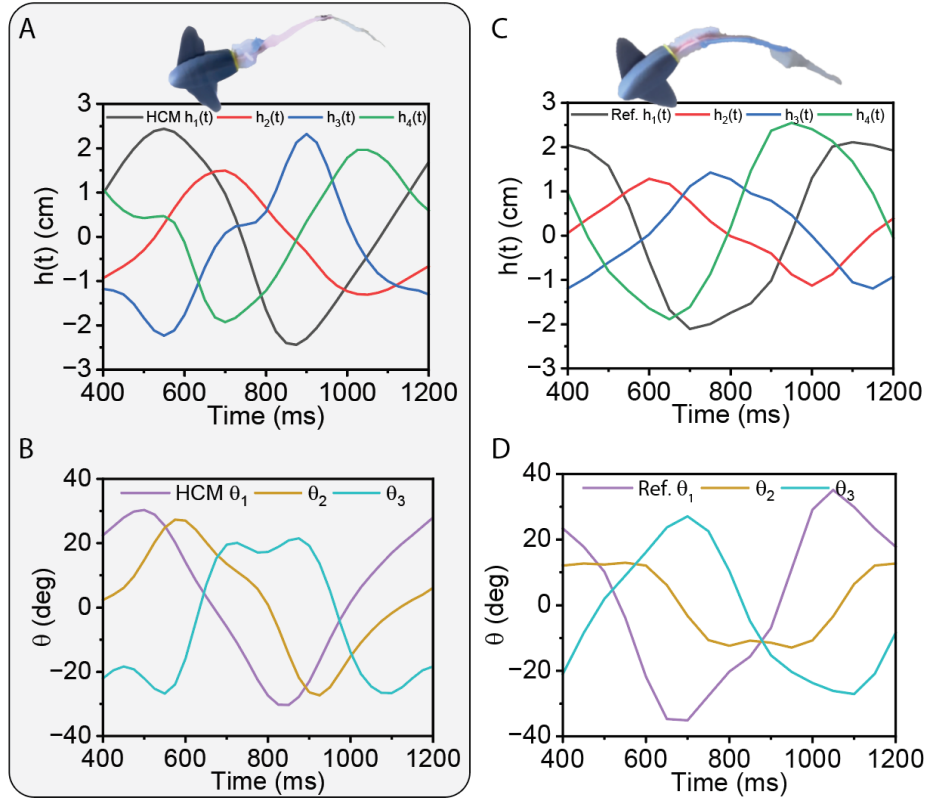

Figure 5 The sampled variable profiles of HCM and reference undulations from the experiments. (A) and (B) The variable plots in a half cycle (~400 ms) of the HCM robot. Nonlinearities like skewness, asymmetry, and cambering-ness are observed in the posterior portion of the HCM fish. (C) and (D) The variables of the reference robot.

The comparison demonstrates that the sine wave has the highest energy efficiency for steady cruising, consistent with natural selection. However, the results also indicate that the sine wave undulation may be the worst in generating thrust and achieving high speeds. Xie et al. [25] and Gao et al. [26] also noticed the better performance of modified sine waves (i.e., cambering sine waves) compared to the ideal sine wave under the same frequency. Xie et al. assumed that the sine wave is adopted by nature over the cambering sine wave because the latter produces a larger recoil and has a lower energy efficiency, indicated by a worse Strouhal Number (SN). Gao et al. achieved a similar conclusion based on the calculation results that the larger thrust of the cambering sine wave is based on an even larger energy input, which corresponds to the results of this work that the cambering sine wave generates more thrust but has a much lower energy efficiency.

On the other hand, fish also use strikes with high velocity and acceleration to generate a large thrust or speed, as is needed during behaviors like hunting and escaping [27]–[29]. The HCM bi-stable undulation generates three times more thrust than the sine wave at the same frequency (1.3 Hz) and 87% of the sine wave efficiency, which shows the HCM undulation's potential for the fish or bionic fish robot during high-speed swimming, especially when HCM simplifies the robotic systems.

Sánchez-Caja et al. [30] proposed that the optimal propulsion solution may lie beyond the scope of living organisms because they have living-body constraints that human inventions are not subject to. Our previous work indicates that [3] the HCM snapping introduces higher stress and strain detrimental to living tissues, which can explain why the HCM undulation is rarely used by nature except in very few cases [31], [32]. However, the novel HCM undulation pattern may be a competitive propelling strategy for aquatic soft robots and bionic underwater vehicles under the abovementioned situations.

IV. OUTLOOK

Due to the lack of comprehensive swimming modeling, the single-link results are still limited and subject to modeling errors and approximations. Therefore, we propose a sampling method for a future multi-link simulation based on the pneumatic fish robots' swimming footage, as shown in Figure 4. Figures 4A and 4B illustrate the configuration evolution of the fish bodies during the HCM and reference swimming, respectively, and the sampled data is shown in Figure 5.

While the characteristic variables on the anterior portion of the HCM fish ($h_1$, $h_2$, and $\theta_1$) show sinusoidal features, the posterior portion ($h_3$, $h_4$, and $\theta_3$) presents nonlinearity like cambering, skewness, and asymmetry. These properties correspond to the working principle of HCMs since they build up elastic energy in their "core" area ($z \leq L_1$, Figure 1A) and transmit and release the energy toward their far end ($z \to L_1$), using the tapering tips as end effectors. On the other hand, the variables of the reference model follow a sine wave pattern. The multi-link scenario can function as a better efficiency comparison of different propulsive patterns and provide baselines for optimizing these locomotion patterns.

V. CONCLUSION

For the first time, we introduce the hair clip mechanism (HCM) undulation resulting from the snap-through buckling of HCMs, a kind of in-plane prestressed mechanism, and demonstrate the higher thrust gained by HCM undulation that we call the aquatic HCM effect. When using an HCM as a robotic propeller of a soft fish robot, the experiment shows a two-fold faster cruising speed (26.54 vs. 13.10 cm/s) than the reference design. A corresponding reduced single-link swinging simulation using Aquarium, a 2D aquatic solver, is conducted to replicate and verify the HCM effect. Results show that the HCM undulation generates 2-3 times more aquatic thrust (16.7 N/m) than the traced reference pattern (6.78 N/m), curve-fit sine pattern (5.34 N/m), and cambering sine pattern (6.36 N/m) and have an energy efficiency 87% of the ideal sine wave. The initial analyses support the assumption that HCM undulation can be a strategy when high-speed swimming or a simpler design is wanted. Meanwhile, a multi-link simulation method is proposed to help verify the effect in the future.

This work on the novel undulation and effect brought by elastic instability may help improve the function of future soft robots, underwater vehicles, and extreme-environment explorers.